\documentclass[letterpaper, 10 pt, journal, twoside]{IEEEtran}  %

\IEEEoverridecommandlockouts                              %

\usepackage{amsmath} %
\usepackage{amssymb}  %
\usepackage{graphicx}
\usepackage{csquotes}
\usepackage[english]{babel}
\usepackage[hidelinks]{hyperref}
\usepackage{balance} %
\usepackage{cite}
\usepackage{soul,color}
\usepackage{float}

\usepackage{booktabs}
\usepackage{multirow}
\usepackage{siunitx}

\addto\captionsenglish{\renewcommand{\figurename}{Fig.}} %

\title{A Laser-guided Interaction Interface for Providing Effective Robot Assistance to People with Upper Limbs Impairments}

\author{Davide Torielli$^{1,2}$, Liana Bertoni$^{1,3}$, Luca Muratore$^{1}$, and Nikos Tsagarakis$^{1}$%
	\thanks{Manuscript received: May, 9, 2024; Revised June, 30, 2024; Accepted June, 30, 2024.}%
	\thanks{This paper was recommended for publication by Editor Jee-Hwan Ryu upon evaluation of the Associate Editor and Reviewers' comments. This work was supported by the European Union's Horizon 2020 research and innovation programme, GA No. 101070292 HARIA.} %
	\thanks{$^{1}$All authors are with HHCM, Istituto Italiano di Tecnologia, Genova, Italy {\tt\small <name.surname>@iit.it}}%
	\thanks{$^{2}$Davide Torielli is also with DIBRIS, University of Genova, Genova, Italy}%
 	\thanks{$^{3}$Liana Bertoni is also with DII, University of Pisa, Pisa, Italy}%
 	\thanks{Digital Object Identifier (DOI): see top of this page.}
}

\begin{document}
	
    \maketitle
    
    \begin{abstract}
Robotics has shown significant potential in assisting people with disabilities to enhance their independence and involvement in daily activities. Indeed, a societal long-term impact is expected in home-care assistance with the deployment of intelligent robotic interfaces. 
This work presents a human-robot interface developed to help people with upper limbs impairments, such as those affected by  stroke injuries, in activities of everyday life. 
The proposed interface leverages on a visual servoing guidance component, which utilizes an inexpensive but effective laser emitter device. By projecting the laser on a surface within the workspace of the robot, the user is able to guide the robotic manipulator to desired locations, to reach, grasp and manipulate objects. 
Considering the targeted users, the laser emitter is worn on the head, enabling to intuitively control the robot motions with head movements that point the laser in the environment, which projection is detected with a neural network based perception module. The interface implements two control modalities: the first allows the user to select specific  locations directly, commanding the robot to reach those points; the second employs a paper keyboard with buttons that can be virtually pressed by pointing the laser at them. These buttons enable a more direct control of the Cartesian velocity of the end-effector and provides additional functionalities such as commanding the action of the gripper. 
The proposed interface is evaluated in a series of manipulation tasks involving a 6DOF assistive robot manipulator equipped with 1DOF beak-like gripper. The two interface modalities are combined to successfully accomplish tasks requiring bimanual capacity that is usually affected in people with upper limbs impairments.
\end{abstract}
    

    \section{Introduction}
\begin{figure}
	\centering
	\includegraphics[width=\linewidth]{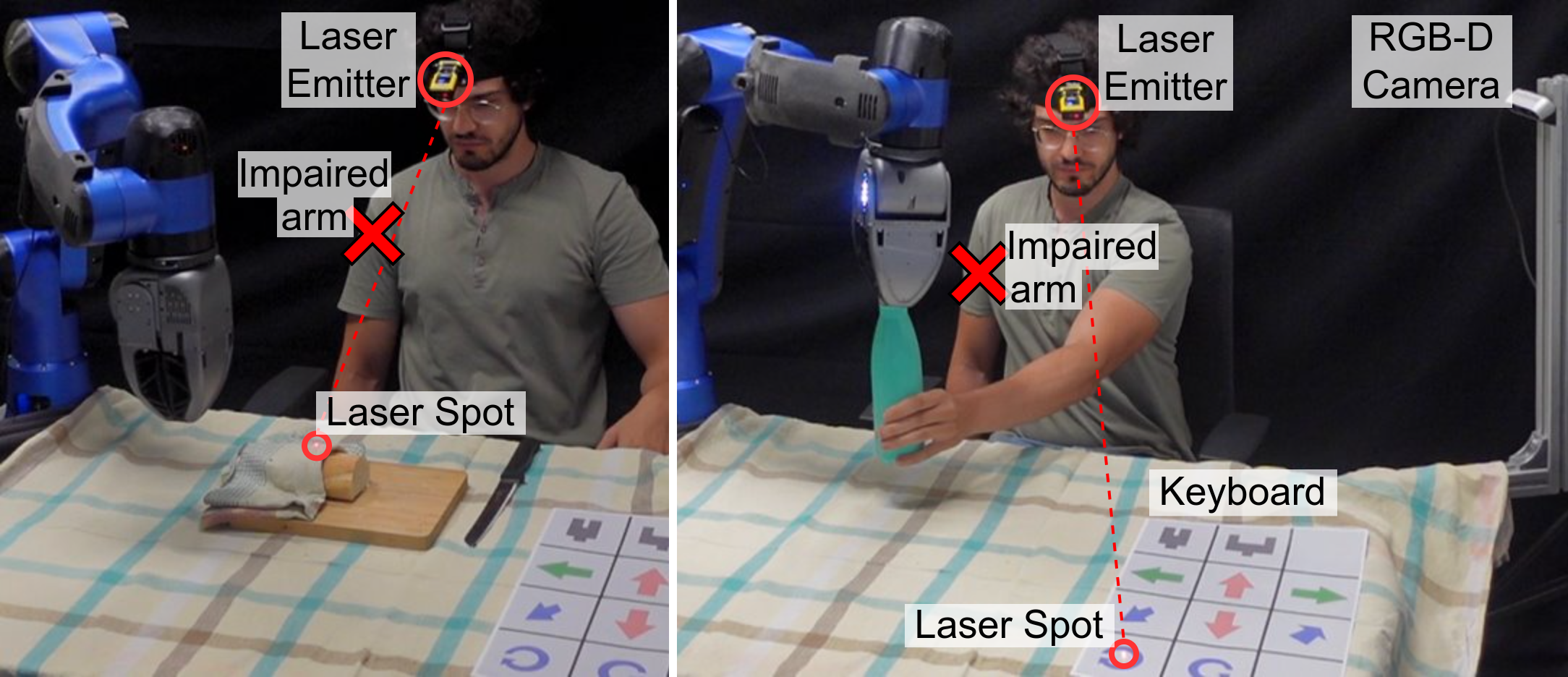}
     \vspace{-15px}
	\caption{The laser-guided interface provides assistance to users with arm impairments. On the left, the user is commanding a target location, to make the robot hold the bread while he will cut it with his healthy arm. On the right, the user is commanding a yaw rotation with the keyboard to unscrew the cap of the bottle.}
	\label{fig:firstPhoto}
    \vspace{-10px}
\end{figure}

\IEEEPARstart{T}{he} recent developments in robotics have demonstrated the potential of intelligent platforms to physically collaborate with humans, providing assistance in various scenarios, like industry~\cite{VILLANI2018}, elderly care~\cite{Bardaro2022}, and Activities of Daily Living (ADL) support~\cite{Petrich2022}. Specifically, research in the domain of ADL assistance has revealed that, by continuously pushing the boundaries of what can be achieved, it is possible to simplify the ADL tasks for individuals with disabilities, ultimately fostering a greater sense of independence and improving their overall quality of life. 

In such a context, this paper presents a laser-guided interaction interface designed to assist users in reaching locations of interest and manipulate objects using a robotic arm. %
The key features/contributions of this work are:

(1) A highly intuitive interface for people with upper limbs impairments is introduced. The interface explores a head-wearable laser pointing device to indicate to the assistive robot the locations of interest related to a task to be executed within the robot workspace. With this interface, head movements in the direction of a specific location of interest result in the laser to point to it. This is a very intuitive way to indicate a target position to the robot end-effector, for example to approach an object to grasp.
 
(2) In addition to commanding the robot to a target position, the interface integrates an additional keyboard control modality. Indeed, the laser can be used to virtually press the buttons of a paper keyboard located in an area of the environment, with each button of the keyboard mapped to gripper actions and specific end-effector directions.
This allows the user to utilize the laser to command the gripper of the robot and the robot itself through Cartesian end-effector velocities. 

(3) The laser point projection is detected by employing a neural-network based vision system that utilizes images from an RGB-D (Red Green Blue-Depth) camera directed at the robot workspace. The detection pipeline is fast enough to allow for a reactive response to laser position changes.

We have conducted a series of experiments demonstrating the efficacy of the interface in a scenario with a custom 6DOF robotic manipulator equipped with a custom gripper (\figurename{}~\ref{fig:firstPhoto}). 
In the experiments, the robot was controlled by healthy users with intentionally limited upper limbs' movement, resembling the motion limitations that impaired people may have.
We have compared the usage of the laser-based assistive interface with a state-of-the-art method based on IMU head tracking. 
Furthermore, we demonstrated how the proposed laser-guided interface allows the user to effectively substitute his/her not functional arm with the assistive robotic manipulator for bimanual co-manipulation ADL tasks, like cutting some bread and opening a bottle (\figurename{}~\ref{fig:firstPhoto}). Furthermore, considering motion impairments on both arms, different users validated the interface with some pick-and-place tasks where the robot is controlled to manipulate alone the objects involved.

\section{Related Works}
A multitude of assistive robotic platforms have been developed in conjunction with sophisticated Human Machine Interfaces (HMIs) to enable individuals with disabilities to control these devices~\cite{Kyrarini2021}.
An example of these platforms is the EDAN platform,
controlled by the user with electromyography (EMG) signals in combination with a head-switch, and enhanced with shared control functionalities~\cite{Vogel2020}. 
One key challenge in developing assistive HMIs is how to effectively utilize the residual motions of impaired users. Interfaces like handheld joysticks~\cite{Wang2012}, and based on arm gestures~\cite{Esposito2021, TPO} may not be suitable for certain disabilities which affect the upper limbs. Similarly, verbal communication~\cite{Poirier2019} may not be feasible due to the user condition. 
To address this challenge, a possibility is to adapt commercial assistive technologies, such as sip-and-puff devices, for controlling the robot~\cite{Jain2019}. However, these devices have very limited control space dimensions (i.e., two inputs associated to sipping and puffing actions), thus requiring the user to frequently switch between several control modes increasing the cognitive workload and slowing down the task execution.
Other approaches employ head gestures tracked by a camera worn on the user's head. For instance, in \cite{Kyrarini2019}, the head gestures are used to navigate a state machine that maps the various manipulator abilities (i.e., Cartesian movements and gripper actions).
Although direct robot control is not always necessary since motions can be learned and autonomously reproduced, the head gestures to navigate the state machine may lack intuitiveness.
Head movements can also be tracked with wearable Inertial Measurements Unit (IMU) devices, and employed to control a cursor to operate a virtual keyboard on a display~\cite{Rudigkeit2020}.
Similarly, eyes movements are utilized to control a virtual cursor on a screen~\cite{Sunny2021} or on an augmented reality device~\cite{Sharma2022}.
Even if these kinds of interfaces have shown their advantages, they may be not intuitive as users have to learn and establish the mappings between their inputs and desired actions.

Other approaches employ laser pointers to select objects of interest in the environment, creating, for example, the \enquote{clickable} world of \cite{Nguyen2008}. This way of interacting with the robot has shown promising results for assistive scenarios, since it is inherently intuitive, and the laser pointer can be comfortably worn on the user head to accommodate people with upper limbs impairments. For example, the interface from \cite{Nguyen2008} has been employed in such scenarios integrating an ear-worn laser pointer to select objects to be picked by a mobile manipulator~\cite{Choi2008}. The laser spot is detected by filtering the laser color wavelength in the images coming from a camera. This approach may require the fine-tuning of filter parameters, and hence can be not robust against different surfaces, or light conditions. 
In \cite{Gualtieri2017}, the laser guides a wheelchair equipped with a manipulator to reach and pick objects. While the laser is pulsating, the spot is detected by looking in a sequence of images for areas with changes in the intensity. This increases the robustness of the detection, but also augment the necessary detection time. 
Another wheelchair system is designed in~\cite{Wilkinson2021}, where a projector illuminates the object selected by the laser, to allow the user to confirm that the proper item has been correctly recognized before grasping it. A second laser is employed to confirm the choice. The two laser spots are detected by looking for two high-value areas in infrared images.
In \cite{Zhong2019}, a neural network is utilized to recognize the pose of objects and to detect the laser projected on them. As before, the objective is to grasp the selected object with the arm of a wheelchair manipulator. The same authors explore a handheld laser pointer flashing it on different objects to command manipulating actions~\cite{Liu2021}.

To facilitate a natural integration of assistive robots into the everyday life of individuals, it is crucial to develop interfaces which incorporate a certain level of robot autonomy, allowing to \textit{share} the control between the human and the robot during the execution of the task. This level of autonomy should be based not only on the specific task requirements but also on the user preferences. Indeed, it has been demonstrated that full robot autonomy is not always the best option~\cite{Kim2012, Bhattacharjee2020}. This consideration has been addressed not only for assistive robotics~\cite{Bustamante2022}, but also for other kind of human-robot collaboration, including workplace settings~\cite{TPO2}.

\begin{figure*}
	\centering
	\includegraphics[width=0.95\linewidth]{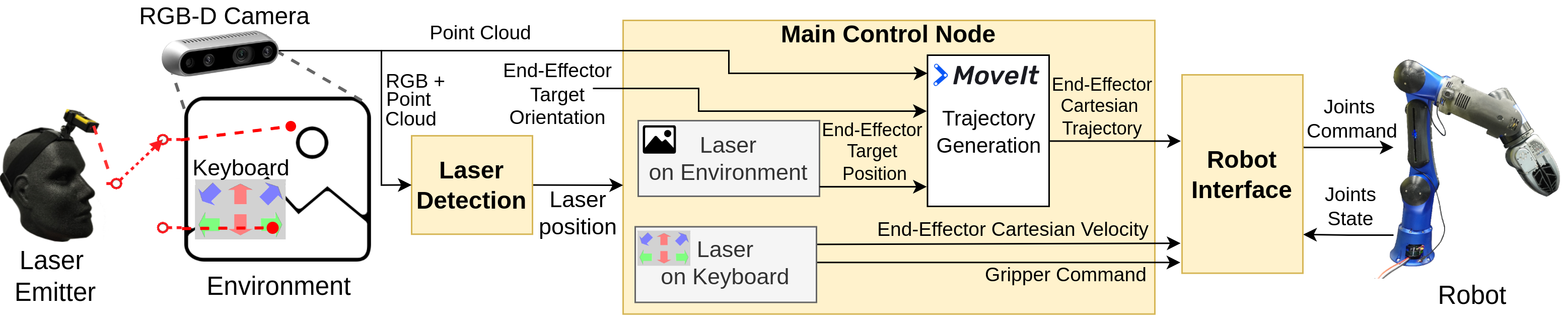}
     \vspace{-5px}
	\caption{Scheme representing the overall interface. Depending on where the laser is pointed (environment or keyboard), the relative control modality is exploited to generate robot commands.}
  	\vspace{-7px}
	\label{fig:controlScheme}
\end{figure*}

    \section{Laser-based assistive interaction interface}\label{sec:system}

We have developed an HMI to assist users with upper limbs motion impairments in reaching location of interests and manipulate objects with a robotic manipulator. Using a lightweight and compact head-mounted laser pointing device, the user can command the robot in an intuitive way without the necessity of any additional input device. The laser spot, projected on the workspace of the robot, is detected by a vision system, which extracts the spot position with respect to the robot. 
The system offers two distinct modalities to interact with the assistive robot based on the detected laser position. 
In the first modality, the robot end-effector is commanded to reach the specific point indicated by the laser. If the point is physically reachable by the manipulator, the system generates and executes a collision-free trajectory to reach the indicated goal. This way of interacting with the robot is very intuitive and natural as it resembles the action of people to naturally orient their head and look toward an object of interest to interact with.
In the second modality, a more direct robot guidance is enabled employing a paper keyboard, placed in a designated area, which buttons can be virtually pressed by projecting the laser onto them. Some buttons are mapped to Cartesian directions, enabling the user to command the robot through Cartesian end-effector velocities, while other buttons give control to the functionalities of the gripper mounted on the arm.

Some interfaces that rely on particular body movements, like head~\cite{Kyrarini2019, Rudigkeit2020} and eyes~\cite{Sunny2021, Sharma2022} necessitate the user to learn and deal with a certain mapping between the particular body movement and the input provided to the robot. Instead, with our interface, the head movements result in pointing the laser in the environment or in the keyboard buttons, intuitively commanding the robot without any prior knowledge of the system. %
Additionally, the user has a real-time visual feedback about the command given to the robot through the perception of the laser spot, without the necessity of any additional communication means like a monitor, an augmented reality device, or some auditory feedback.
The advantages of laser-based approaches have been shown in the mentioned previous works. In these cases, usually, the laser was used to select an object to grasp~\cite{Nguyen2008, Choi2008, Gualtieri2017, Wilkinson2021, Zhong2019} or to manipulate~\cite{Liu2021}.  
Differently, the functionality of our interface does not focus on the selection of an object considering its automatic grasping. %
Instead, the interface aims to facilitate the regulation of the robot motion and interaction toward any point of interest, still maintaining the intuitiveness of the laser guidance approach.
By combining seamlessly the two control modalities, we provide a combination of robot autonomy and low level Cartesian control that provide flexibility for various tasks. 
This allows not only to face grasping tasks, but also to perform co-manipulation tasks with the robot, to replace the functionality of the impaired limb. For example, the robot can be guided to hold the bread in a point chosen by the user, such that he/she has enough space to cut it (\figurename{}~\ref{fig:firstPhoto}). Furthermore, allowing the control of the end-effector after the grasping, the user can move the object where he/she prefers, or it is more convenient, like near his/her healthy arm to open a bottle (\figurename{}~\ref{fig:firstPhoto}). In addition, as demonstrated in other experiments, pick-and-place tasks can be still achieved even without an automatic grasping system utilized in other works.
Compared to the previous works that made use of a laser based interface, apart from the added flexibility in the control of the robot motion, providing such functionalities greatly increases the interaction of the human with the robot. This augmentation enhances the sense of involvement of the human subject in the regulation of the execution of task. Hence, impaired users can play an active part in the task, which is important to help them regain their sense of autonomy~\cite{Kim2012, Bhattacharjee2020}.
Another advantage of the proposed interface is that no additional input is required to command manipulation actions on the object since the user can manipulate it by commanding the robot with the paper keyboard using the same laser pointer device.
This relaxes the need for employing additional devices that may require the engagement of the healthy upper limb (e.g.\ pressing additional buttons, making gestures, etc.).
For example, in \cite{Liu2023} the authors show how flashing the laser can be employed to command different manipulation actions, but this requires switching on and off the laser, hence employing the user hand.
Indeed, the user's arm should be left free from other motion activities in order to be fully available to collaborate with the robot arm in bimanual tasks. Other interfaces that do not require engaging the healthy upper limb, like vocal recognition interfaces, would complicate the system, may lack robustness especially in noisy environments, and, in some cases, are not feasible because of users' speech impairments. 

In conclusion, the two modalities well combine to provide a variety of possibilities, from commanding a target location, to controlling the full end-effector pose and activating the gripper. The keyboard modality also allows reaching locations that cannot be indicated by the laser (e.g.\ because of occlusions or the absence of a surface to project the laser), and to account for any potential errors in the end-effector pose achieved with the other modality. In the end, recognizing also the importance of offering different levels of autonomy~\cite{Kim2012, Bhattacharjee2020}, the interface permits the user to choose the best control modality based on the task and their preferences, switching between them simply by adjusting the laser position.

    \section{System Architecture}\label{sec:methods}
\begin{figure*}
	\centering
	\includegraphics[width=0.88\linewidth]{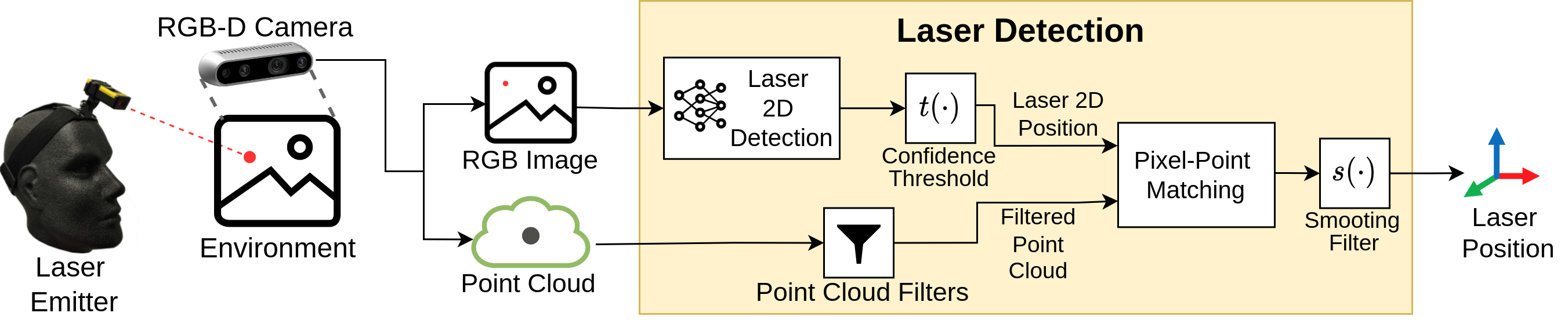}
	\caption{Scheme representing the laser detection pipeline. The RGB-D camera provides RGB images and point clouds. From the RGB image, the neural network extracts the 2D pixel coordinates of the laser spot. These coordinates are then matched with the point cloud to extract the 3D position of the laser.}
	\vspace{-5px}
	\label{fig:visionScheme}
\end{figure*}

\begin{figure}
	\centering
	\includegraphics[width=1\linewidth]{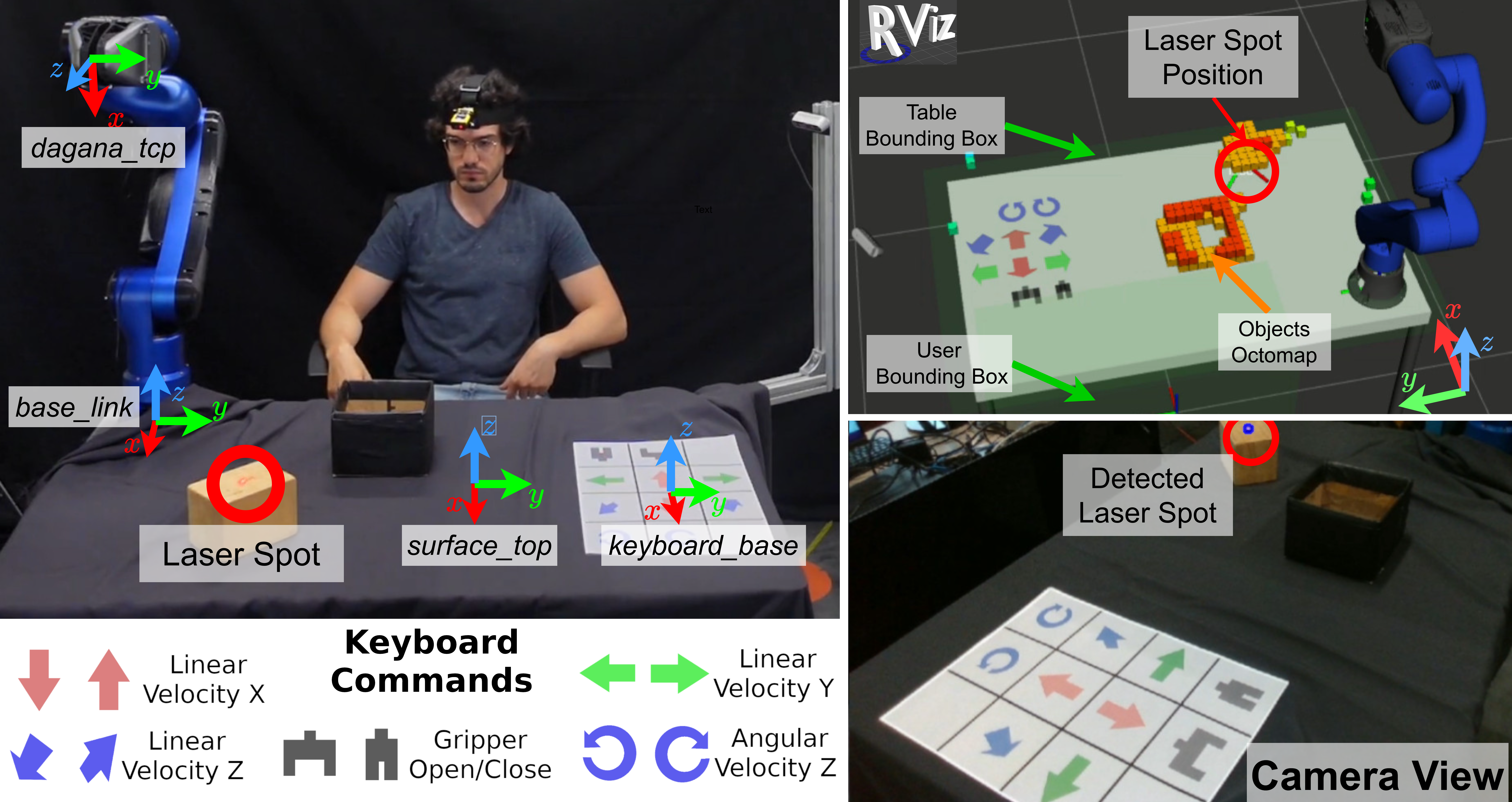}
	\vspace{-13px}
	\caption{Experimental setup. On the left, the user points the laser to a goal. On the top-right, RViz displays the laser spot and the obstacles considered for generating the trajectory to the goal. On the bottom-right, the laser spot is detected by the system.}
	\vspace{-10px}
	\label{fig:setup_with_frames}
\end{figure}

An overall scheme of the system is shown in \figurename{}~\ref{fig:controlScheme}. Hereafter, we describe each component in detail.

\subsection{Laser Detection}
The laser spot detection pipeline is schematized in \figurename{}~\ref{fig:visionScheme}. An RGB-D camera provides RGB images and depth images (i.e., a point cloud) aligned to each other. In the upper branch of the \textit{Laser Detection} module, a neural network based on YOLOv5~\cite{yolov5} is exploited to extract the 2D pixel coordinates of the laser spot. The network has been fine-tuned with a custom dataset of $385$ images which contain the laser projection on different surfaces. 
The use of the neural network is robust enough to detect the laser spot projected on objects of different materials. In contrast, other pipelines that rely on traditional computer vision algorithms, like blob detection, may require heavy filtering and fine-tuning of parameters, depending on the light conditions and surfaces where the laser is projected. Despite employing a more complex neural network in our interface, the inference of the 2D pixel coordinates of the laser projection is fast enough to reach the frequency at which the images are captured by the camera (\SI{30}{\hertz}). This enables real-time responsiveness in detecting changes of the laser spot position which permits to guide the robot without noticeable delays.
In the lower branch of the scheme, with a cascade of point cloud filters, we eliminate points which are outside the robot workspace%
, and that correspond to the robot body.
In the \textit{Pixel-Point Matching} block, the inferred 2D pixel coordinates of the laser projection are used as indexes to access the corresponding 3D point of the cloud. This is an operation that requires no significant amount of time, since the RGB image and the point cloud are aligned. %
At the end of the scheme, a smoothing filter removes the high-frequency fluctuations in the laser position, providing a more stable reference. 
To calibrate the camera position with respect to the robot, in a preliminary phase we have utilized an \textit{ArUco} marker\footnote{\url{http://wiki.ros.org/aruco_detect}} fixed in a known position (i.e., in the \textit{surface\_top} frame of \figurename{}~\ref{fig:setup_with_frames}). The laser spot detection stack is available at \url{https://github.com/ADVRHumanoids/nn_laser_spot_tracking}

\subsection{Main Control Node and Robot Interface}
As shown in the scheme of \figurename{}~\ref{fig:controlScheme}, based on the position of the laser projection, one of the two control modalities is exploited, generating different commands for the robot.

\subsubsection{Environment Control Mode}
When the user points a generic location in the robot's workspace, the system interprets the laser projection as an end-effector target position $\boldsymbol{x} \in \mathbb{R}^3$. To avoid selecting unwanted locations, the user must keep the laser sufficiently still in a position for a certain amount of time. 
In the experiments, the end-effector target orientation is set in such a way to orient the gripper in a pose suitable to grasp objects from the top. Future works will include the automatic computation of the target orientation depending on the shape and position of the pointed object.
From the end-effector target pose, a Cartesian trajectory is computed with MoveIt~\cite{moveit}, avoiding singularities, self-collisions, collisions with manually added obstacles (like the \textit{Table} and \textit{User Bounding Boxes} of \figurename{}~\ref{fig:setup_with_frames}), and obstacles detected by the same camera used to detect the laser (\textit{Objects Octomap} of \figurename{}~\ref{fig:setup_with_frames}).

\subsubsection{Keyboard Control Mode}
This mode offers a more direct control of the robot, since the user, by means of the buttons of the keyboard, can command end-effector Cartesian velocities and gripper actions. 
In our setup (\figurename{}~\ref{fig:setup_with_frames}), the keyboard has six buttons to command the linear Cartesian velocities (positive and negative verse for each of the three axis $x$, $y$, $z$), two buttons to command the angular Cartesian velocities along the $z$ axis of the end-effector (positive and negative), and two buttons to open/close the gripper. 
The linear velocities are relative to the reference frame \textit{base\_link} shown in \figurename{}~\ref{fig:setup_with_frames}, with the $x$-axis along the width of the table, the $y$ along the length, and the $z$ perpendicular to its surface, pointing up. For the angular velocities, the $z$-axis is the one pointing out from the gripper (\textit{dagana\_tcp} frame).
A button is virtually pressed when the laser spot is detected in its specific area. Each button of the keyboard has a known size (\SI{0.105}{\meter} x \SI{0.099}{\meter}), so its area's position with respect to the keyboard center (\textit{keyboard\_base} frame) is known, and the keyboard itself is fixed in a certain position with respect to the robot. Similarly to the environment control mode, a button is considered pressed after the laser is kept in its area for a certain amount of time.
When the user selects a keyboard button related to the end-effector velocities, the wanted direction and verse is commanded with a certain magnitude, while the gripper is commanded in a discrete open/close mode.
The buttons' configuration, sizes and positions have been chosen according to the tasks' needs. Indeed, multiple choices can be made, for example adding buttons to control other Cartesian angular directions, or to change the speed of the robot.

\subsubsection{Robot Interface}
At the rightmost part of \figurename{}~\ref{fig:controlScheme}, the Robot Interface computes the inverse kinematic and communicates with the robot, employing the CartesI/O~\cite{cartesio} control framework and the XBot2~\cite{XBot2} middleware.

    \section{Experimental Validation}\label{sec:exp}
We have validated our interface with a series of ADL tasks where healthy users\footnote{The experimental protocol (HARIA [SCEN 1] 12/2023) was approved by the ethics committee CAREUS. Participants accepted informed consent.}, simulating different upper limbs impairments, collaborate with a 6DOF manipulator equipped with a 1DOF gripper. 
The camera installed in the robot workspace scene, providing the images to track the laser spot and to detect possible obstacles on the table, was an Intel Realsense D435 RGB-D camera. With the environment control mode, the system accepts a goal if the laser spot is kept in position for $\SI{3}{\second}$ with a $\SI{0.04}{\meter}$ radius tolerance.
For the keyboard control mode, the time has been reduced to $\SI{1}{\second}$. This reduced duration has been chosen because of the placement of the keyboard aside, thus having a lower likelihood of accidentally pointing the laser in this region. The magnitude of the linear and angular velocities commanded with the keyboard have been set to $\SI{0.025}{\meter/\second}$ and $\SI{0.25}{\radian/\second}$, respectively.

\subsection{Comparison of Laser-based vs IMU-based Interfaces}
\begin{figure}[H]
	\begin{minipage}[s]{.7\linewidth}
		\includegraphics[width=\linewidth]{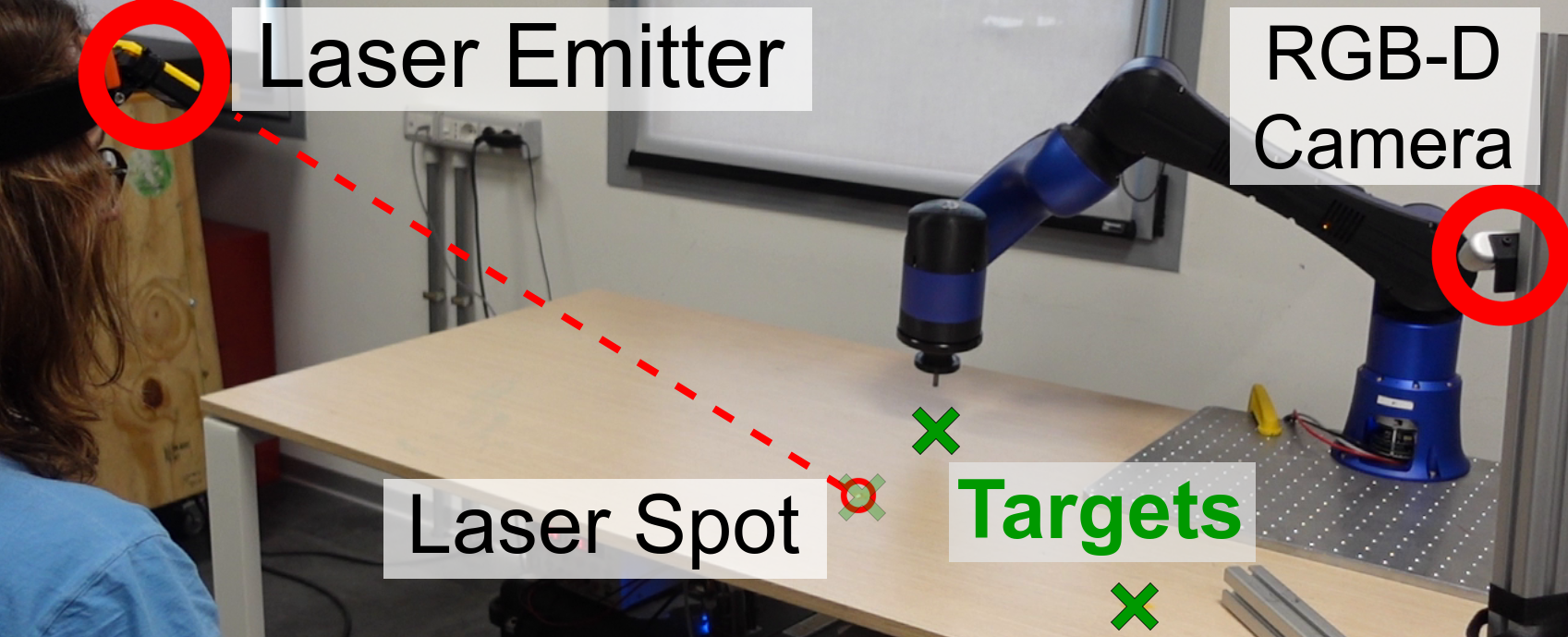}
		\vspace{5px}
	\end{minipage}
	\begin{minipage}[s]{.28\linewidth}%
		\vspace{-23px}
		\begingroup
		\setlength{\tabcolsep}{2pt} %
		\begin{tabular}{lc}
			& {\scriptsize \textbf{Laser}} \\
			\toprule
			{\scriptsize Time[s]}  & {\scriptsize 26.9$\pm$7.1}  \\
			{\scriptsize Err.[mm]} & {\scriptsize 30$\pm$5} \\  
			{\scriptsize Mov.[rad]} & {\scriptsize 1.77$\pm$0.84} \\                  
			\bottomrule                                             
		\end{tabular}
		\endgroup

	\end{minipage}

	\begin{minipage}[s]{.7\linewidth}
		\vspace{-12px}
		\includegraphics[width=\linewidth]{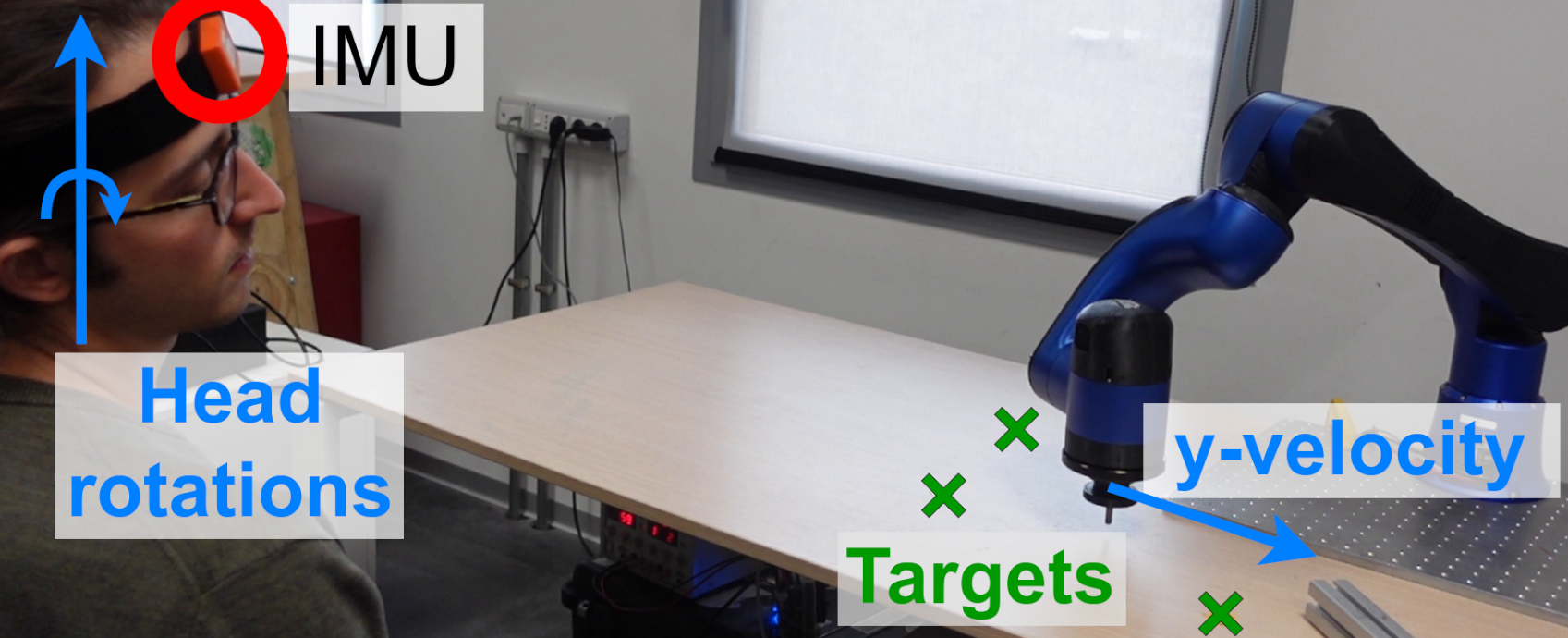}
	\end{minipage}
	\begin{minipage}[s]{.28\linewidth}%
		\vspace{-18px}
		\begingroup
		\setlength{\tabcolsep}{2pt} %
		\begin{tabular}{lc}
			& {\scriptsize \textbf{IMU}} \\
			\toprule
			{\scriptsize Time[s]}  & {\scriptsize 62.5$\pm$13.8} \\
			{\scriptsize Err.[mm]} & {\scriptsize 49$\pm$10} \\  
			{\scriptsize Mov.[rad]} & {\scriptsize 17.86$\pm$7.54} \\                  
			\bottomrule                                             
		\end{tabular}
		\endgroup
	\end{minipage}
	\caption{Our laser-based interface is compared with a IMU-based interface in a reaching targets task.}
	\label{fig:comparison}
\end{figure}
In this comparison, the task requires the user to command the robot's end-effector toward three different locations placed on the table (\figurename{}~\ref{fig:comparison}). We have compared our laser-based interface with another one based on common state-of-the-art interfaces which employ IMU sensors~\cite{Rudigkeit2020}. With the latter, the user controls the robot through head movements tracked by an IMU device~\cite{xsensDriver}, in our case worn on the forehead. At each instant, the user can control the end-effector in a 2DOF plane: $z-y$ or $x-y$. Head rotations command always the $y$ Cartesian linear velocities (\figurename{}~\ref{fig:comparison}). Instead, with a lateral flexion gesture of the head, kept in position for one second, the control plane can be changed from $z-y$ to $x-y$ and vice-versa, thus mapping the head flexion/extension to the $z$- or $x$-velocities.
The displacement of the head is multiplied by a gain to compute the velocity command. Such a gain is set to have maximum velocities of the end-effector comparable with our laser-based interface (in the environment control mode), but also to not generate too high velocities that would make very difficult to stop the end-effector at the required locations. 
Four participants executed the task three times with each interface. Even with the laser interface, users wore an IMU on the forehead to compare the total head movements during the task. In this simple scenario, the keyboard control mode of our interface was not available.
Results are shown in the tables of \figurename{}~\ref{fig:comparison}, where the mean data of all the trials of all the subjects is considered.
\textit{Time} is the time necessary to complete the task. \textit{Err.} is the distance between where the end-effector is stopped and the relative target. \textit{Mov.} measures the total head movements during the task: mathematically, it is the numeric integration of the norm of the angular velocity collected by the IMU.   
The reported data, and the recordings shown in the attached video, demonstrate how the use of the laser-based interface results in faster, more precise, and less fatiguing executions, thanks to the intuitiveness of commanding the robot by directing the head toward the targets.

\begin{figure*}
	\centering
	\includegraphics[width=0.24\linewidth]{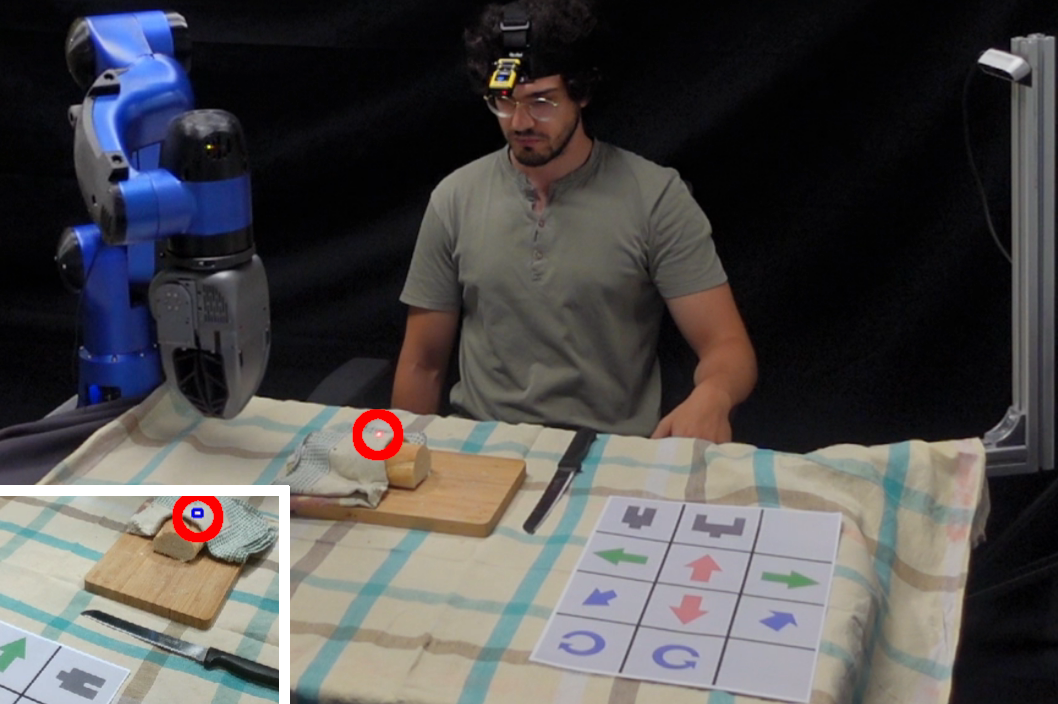}
	\includegraphics[width=0.24\linewidth]{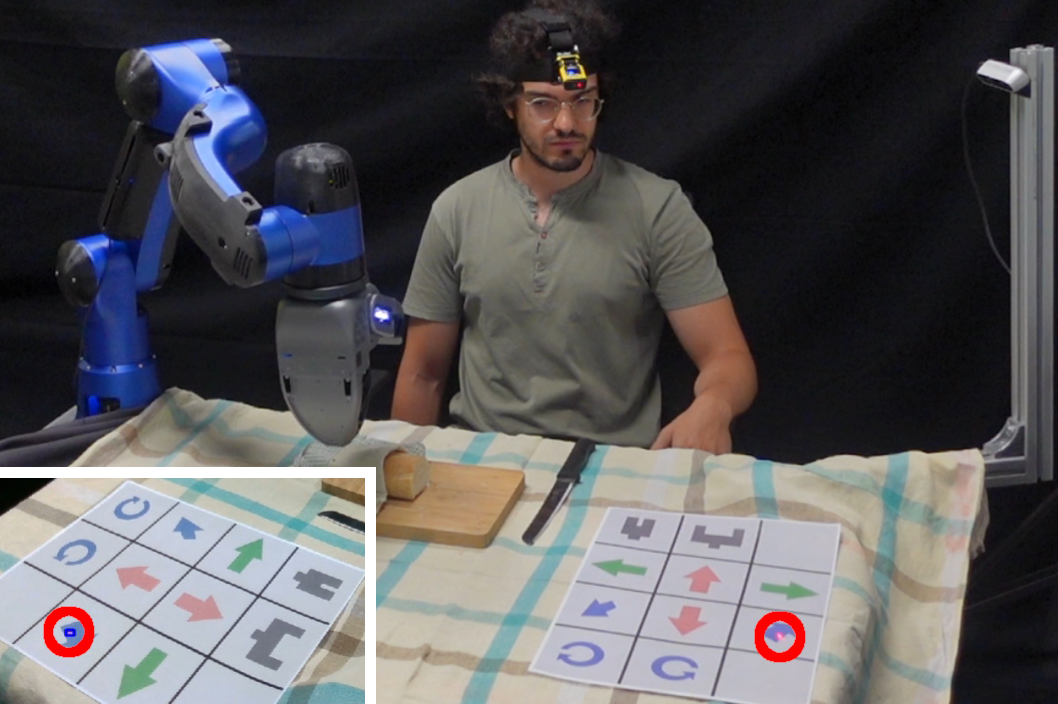}
	\includegraphics[width=0.24\linewidth]{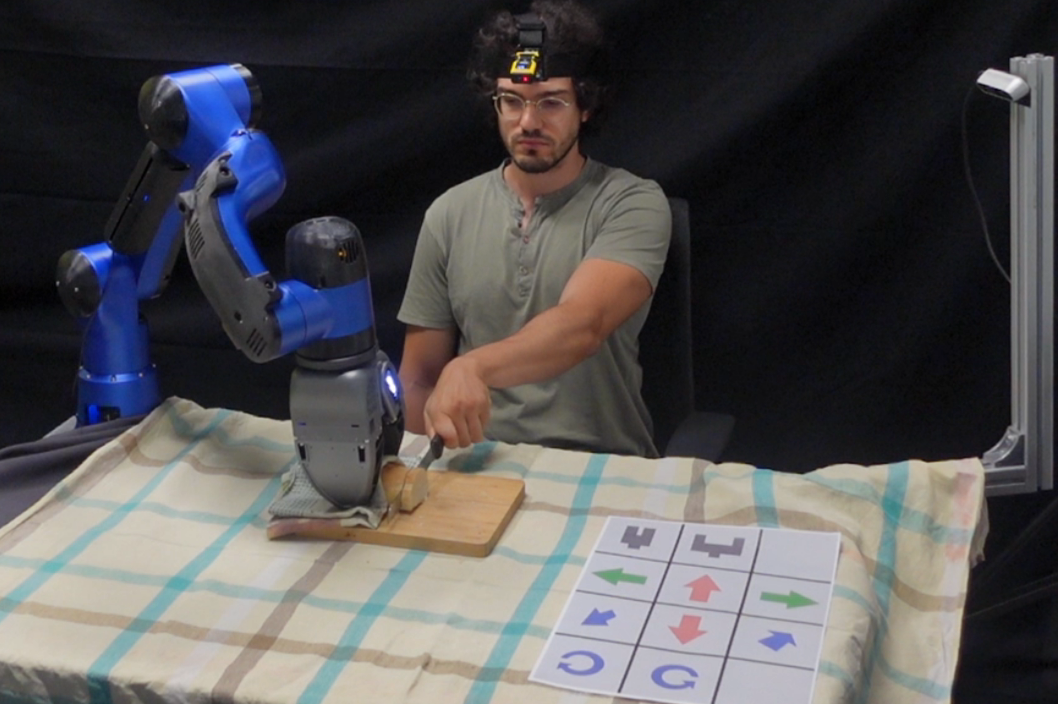}
	\includegraphics[width=0.24\linewidth]{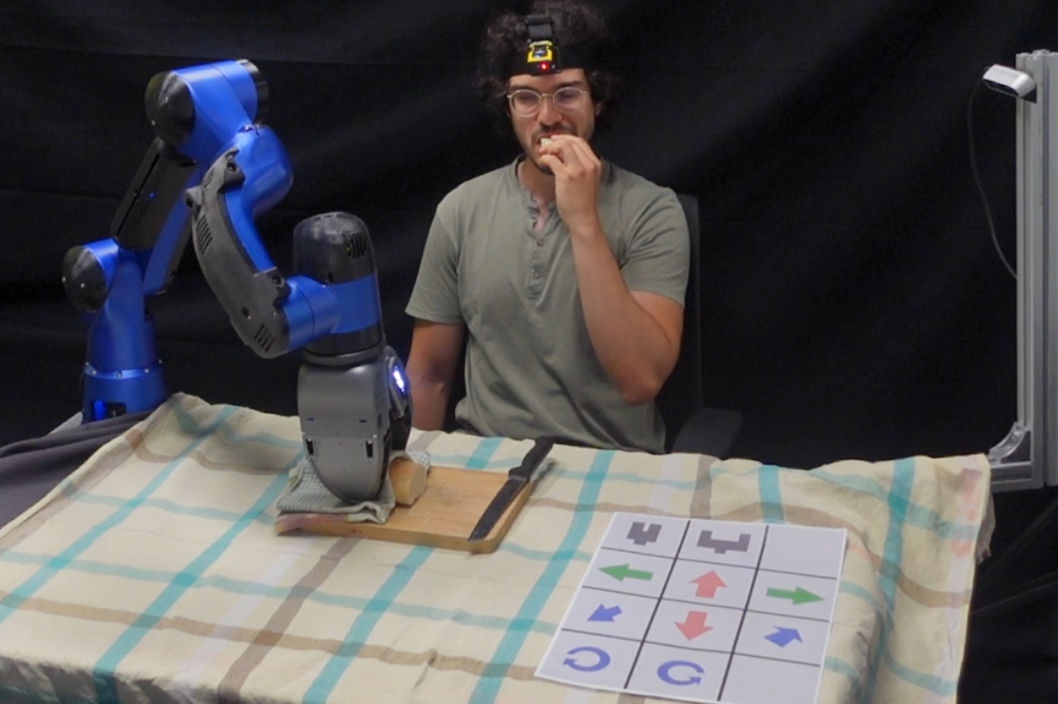}
	\vspace{-5px}	
	\caption{The \enquote{cutting bread} experiment, with the camera view added in the frames when the robot is commanded, showing the detected laser spot.}
	\label{fig:pane3-frames}
\end{figure*}

 \begin{figure*}
 	\centering
 	\includegraphics[width=0.192\linewidth]{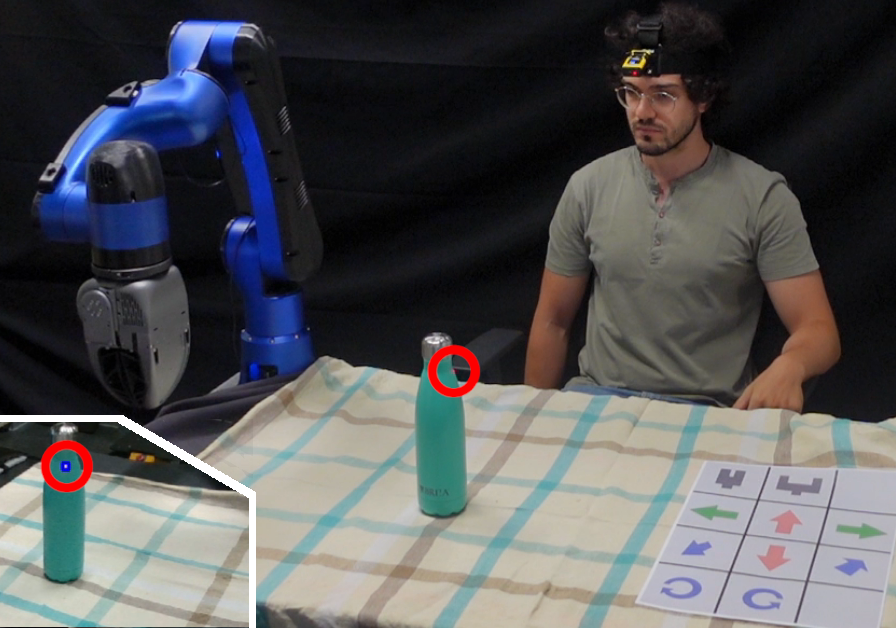}
 	\includegraphics[width=0.192\linewidth]{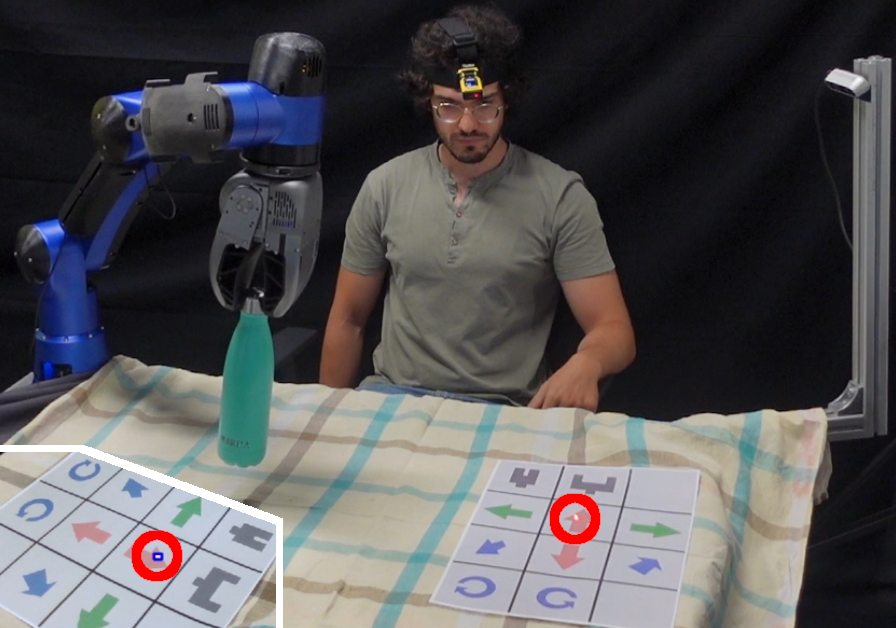}
 	\includegraphics[width=0.192\linewidth]{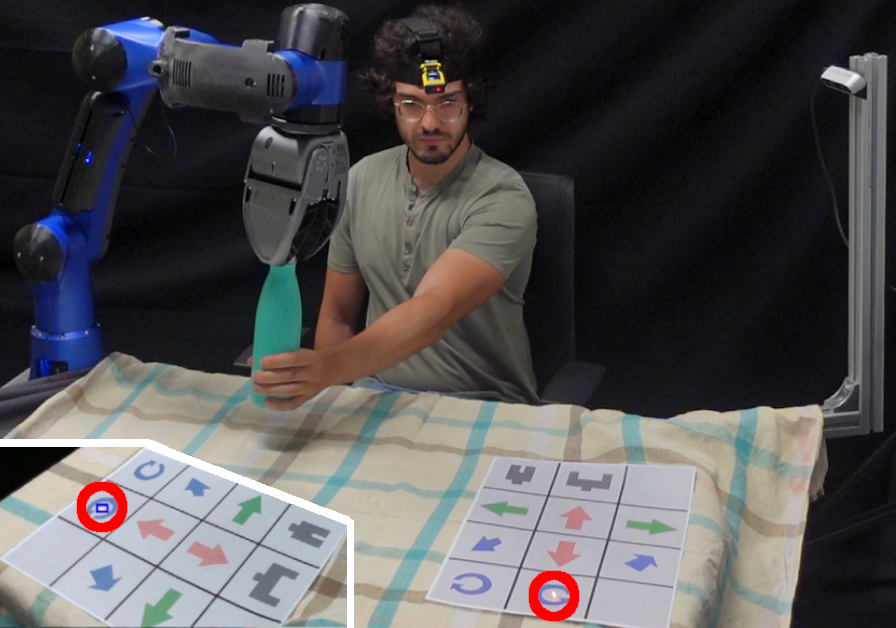}
 	\includegraphics[width=0.192\linewidth]{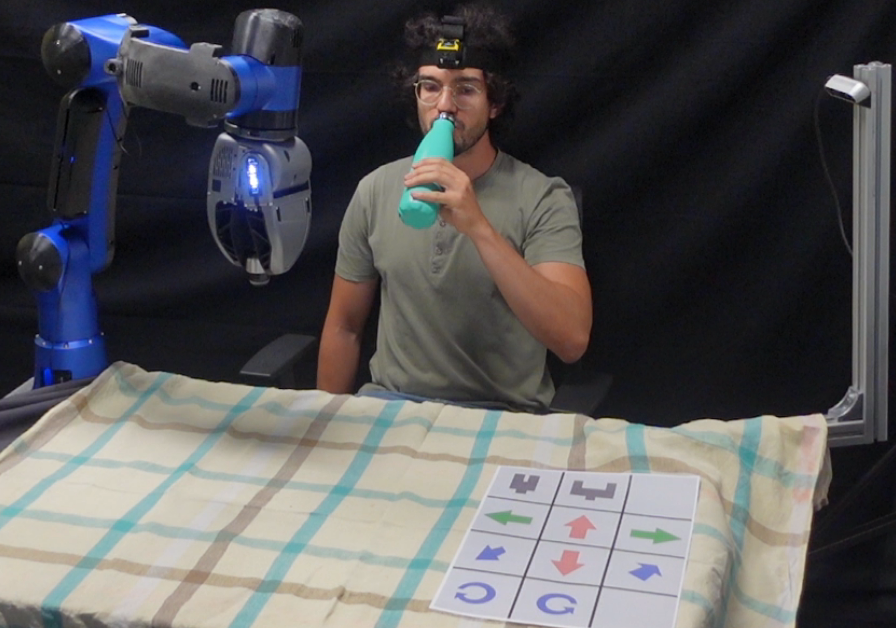}
 	\includegraphics[width=0.192\linewidth]{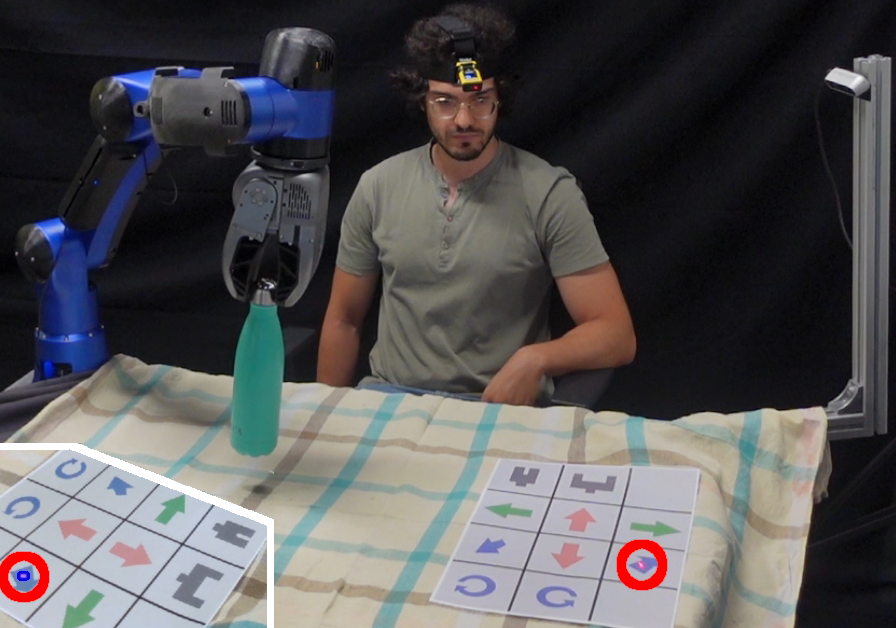}
	\vspace{-5px}	
 	\caption{The \enquote{bottle} experiment, with the camera view added in the frames when the robot is commanded, showing the detected laser spot.}
 	\label{fig:bottiglia-frames}
	\vspace{-10px}
 \end{figure*}

\begin{figure}
	\centering
	\includegraphics[width=1\linewidth]{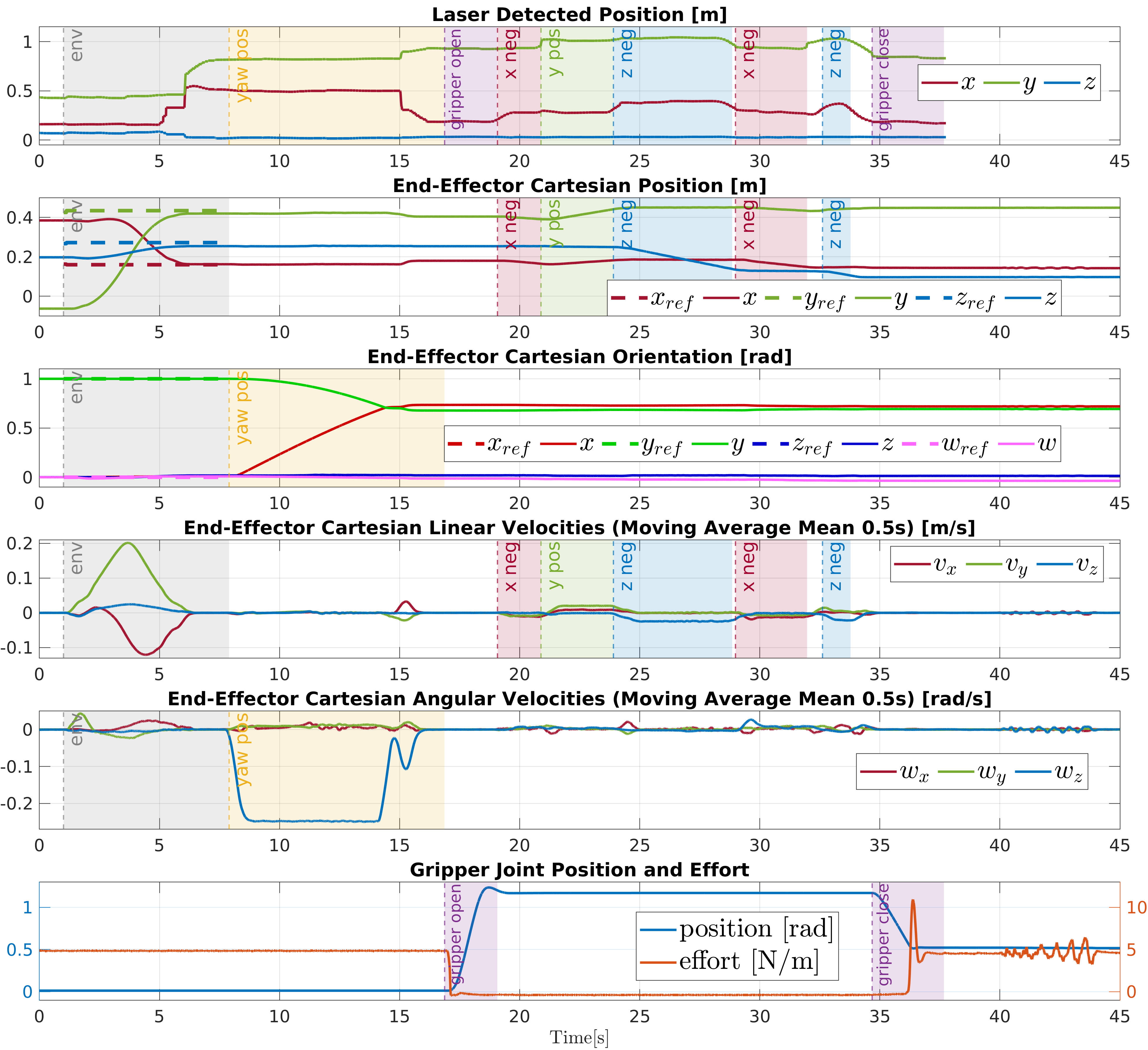}
	\vspace{-14px}
	\caption{\enquote{Cutting bread} experiment plots, highlighting the intervals when the user is commanding the robot with a specific modality.}
	\label{fig:pane3-plot-all}
	\vspace{-10px}
\end{figure}

 \begin{figure}
 	\centering
 	\includegraphics[width=1\linewidth]{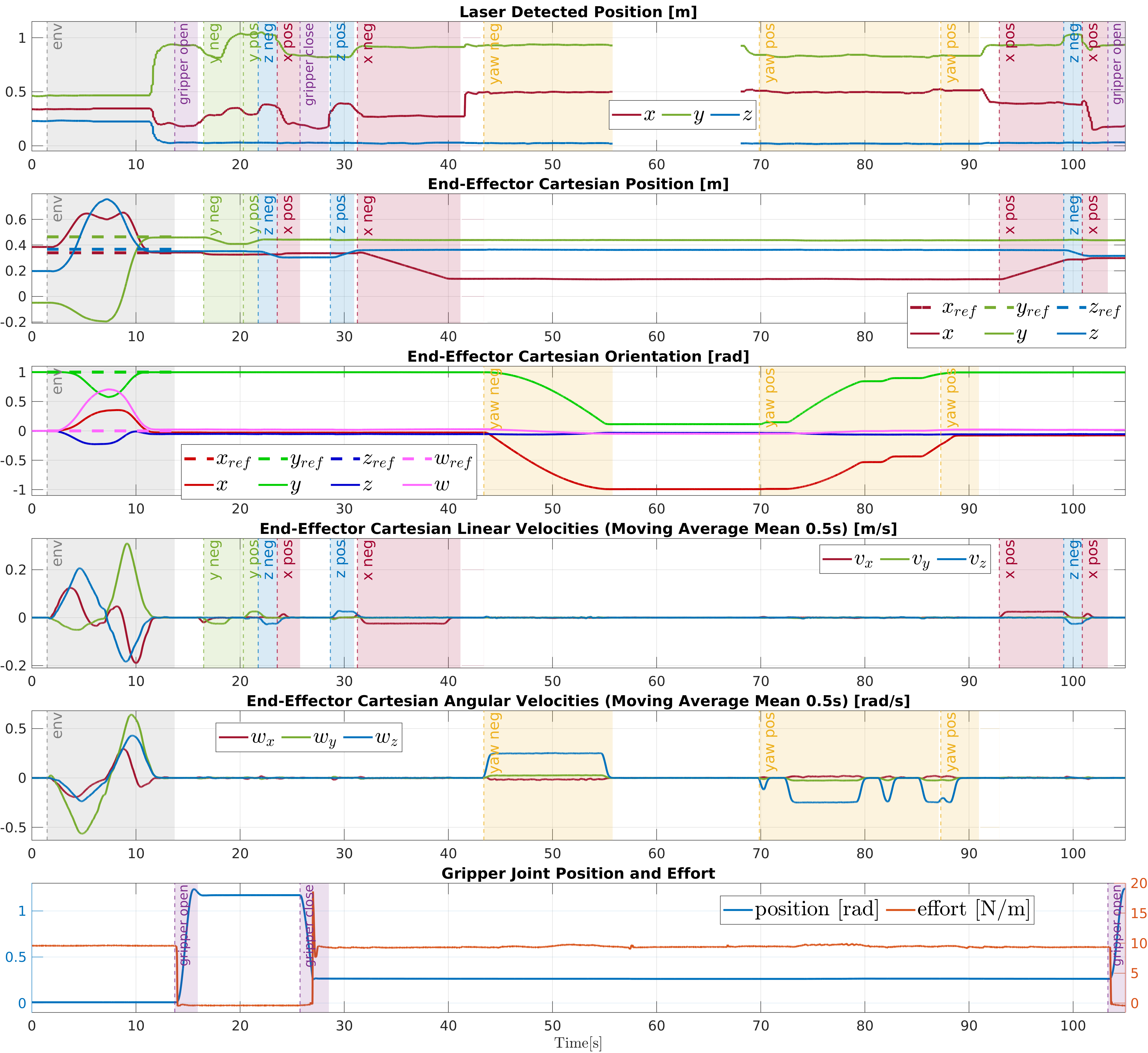}
 	\vspace{-14px}
 	\caption{\enquote{Bottle} experiment plots, highlighting the intervals when the user is commanding the robot with a specific modality.}
 	\label{fig:bottiglia-plot-all}
 		\vspace{-10px}
 \end{figure}

\begin{figure*}
	\centering
	\includegraphics[width=0.24\linewidth]{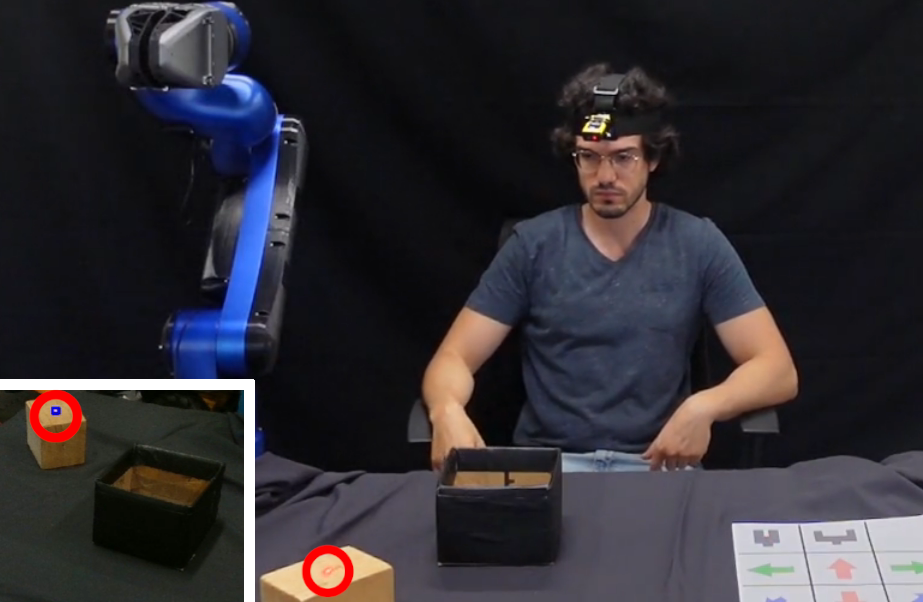}	\includegraphics[width=0.24\linewidth]{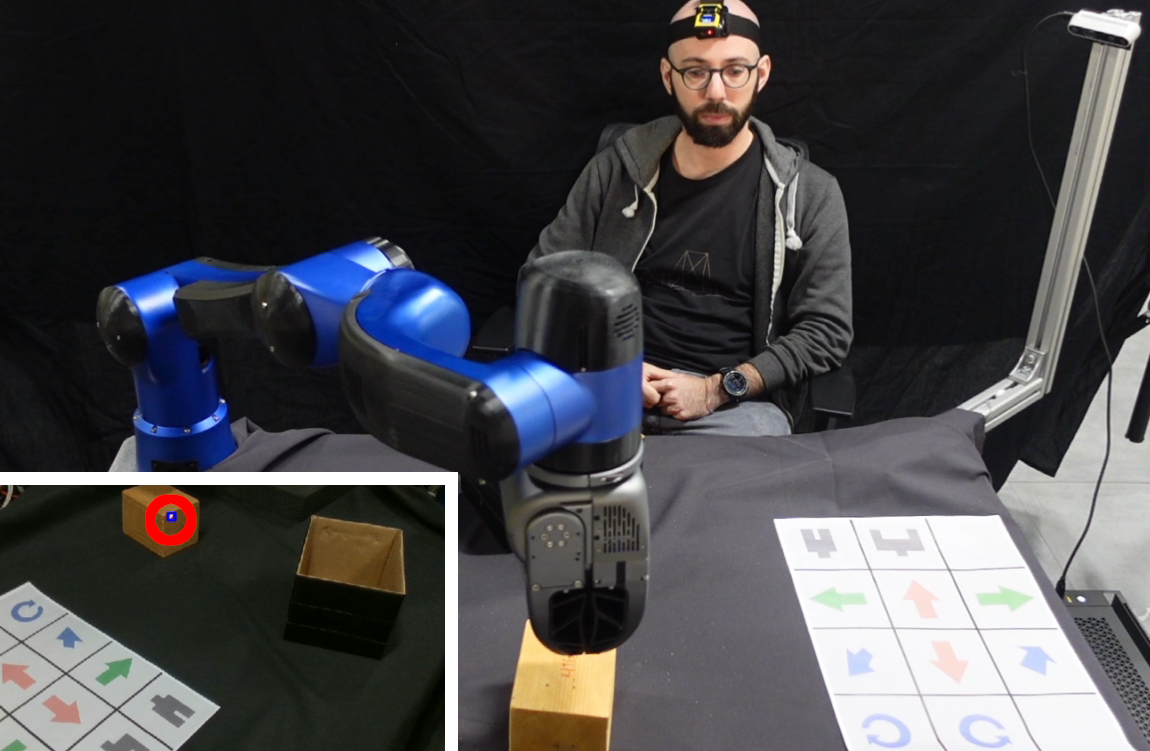}
	\includegraphics[width=0.24\linewidth]{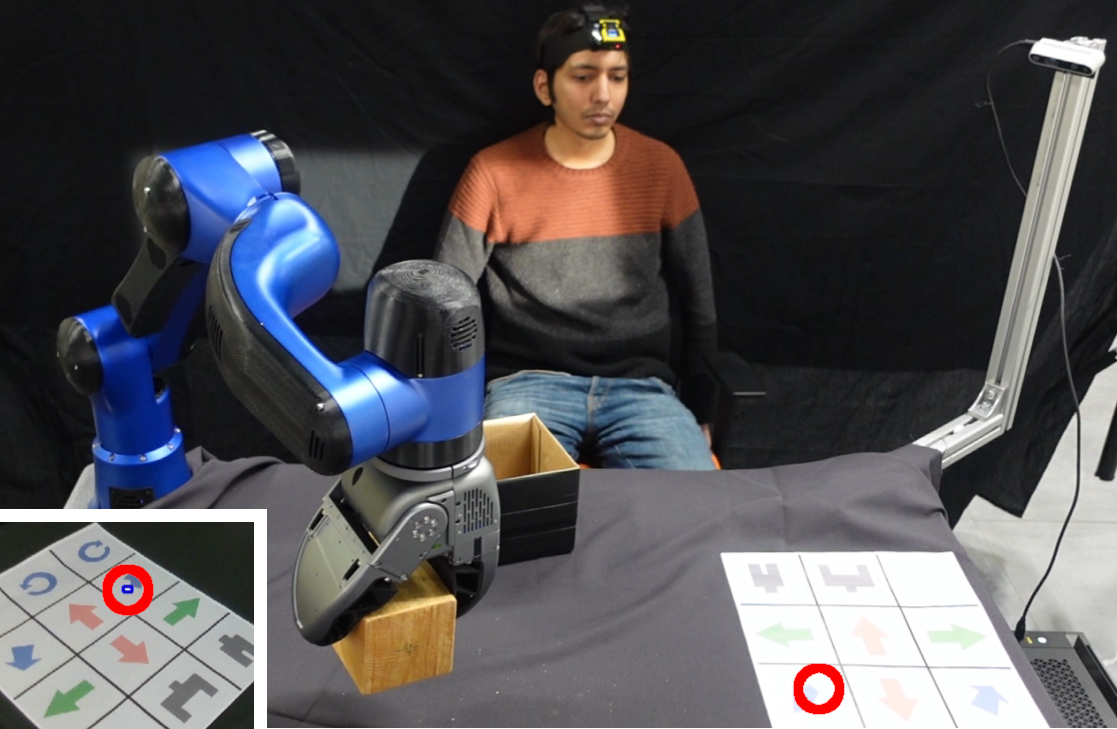}
	\includegraphics[width=0.24\linewidth]{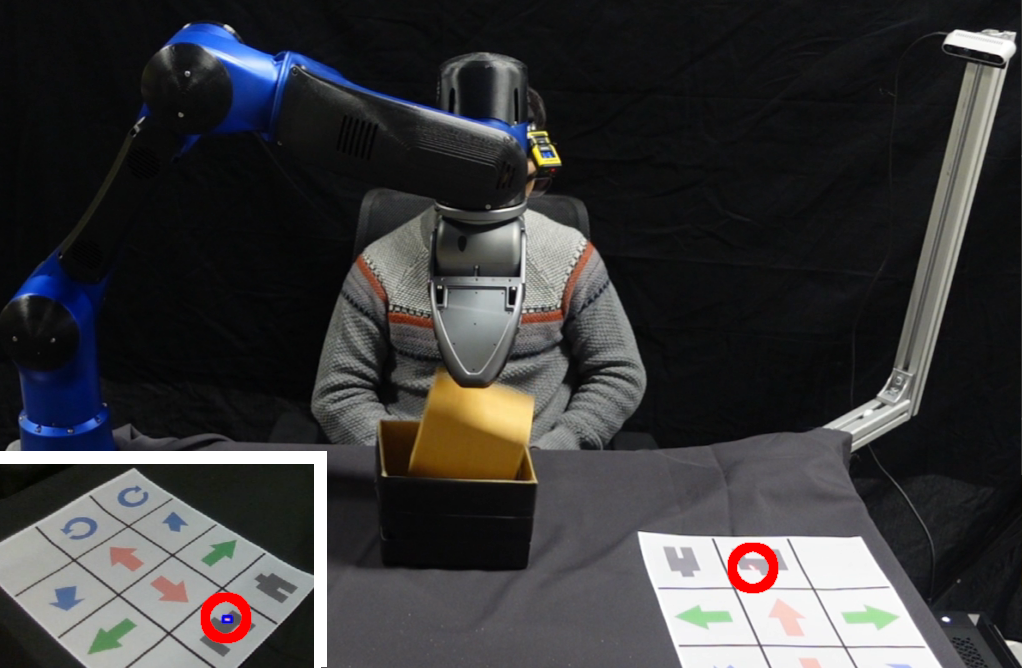}
	\vspace{-5px}
	\caption{The \enquote{wooden block} pick-and-place experiment, executed by different users.}
	\label{fig:exp2-frames}
\end{figure*}

\begin{figure*}
	\centering
	\includegraphics[width=0.24\linewidth]{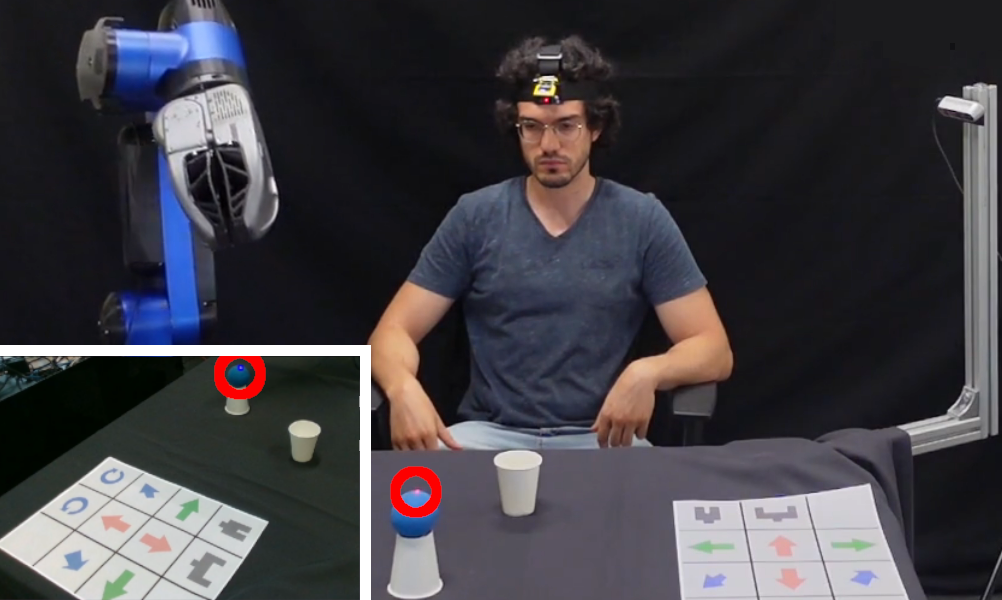}
	\includegraphics[width=0.24\linewidth]{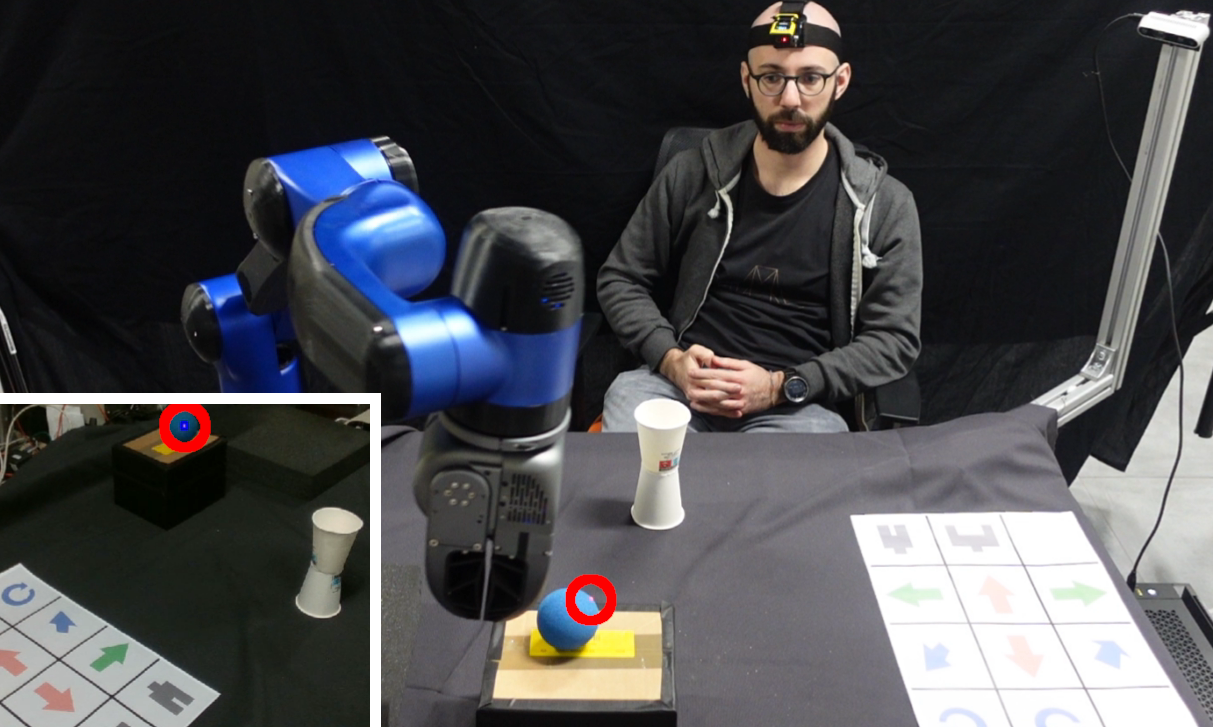}
	\includegraphics[width=0.24\linewidth]{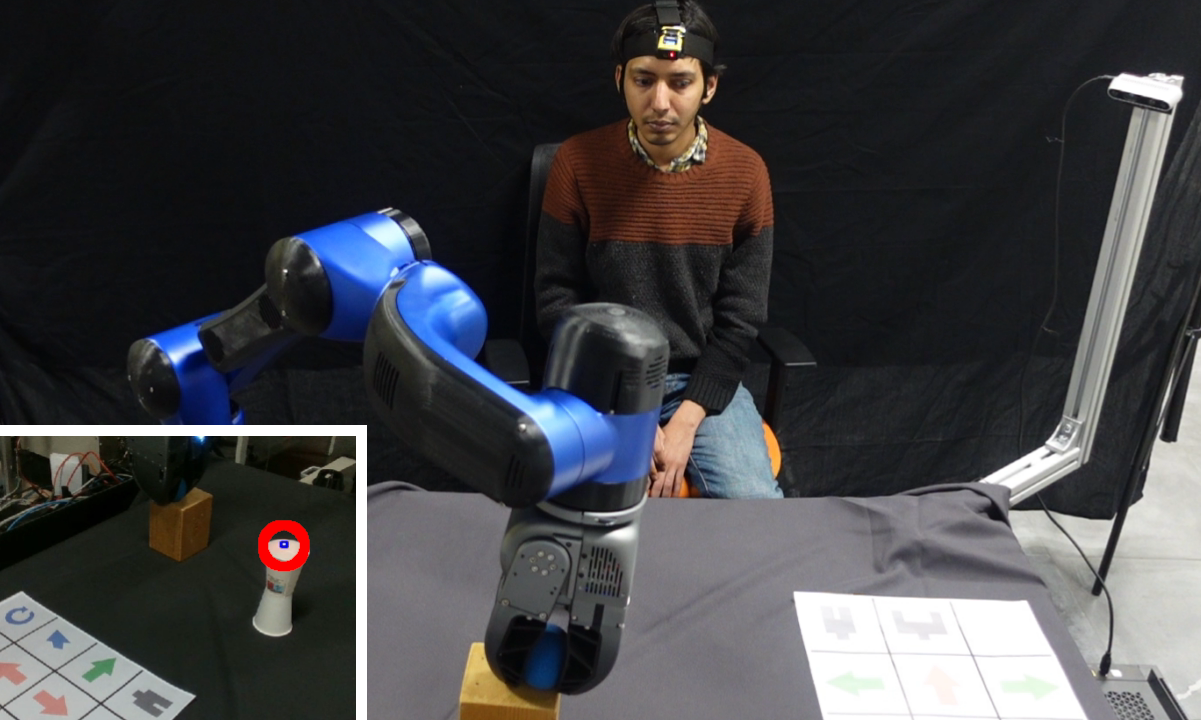}
	\includegraphics[width=0.24\linewidth]{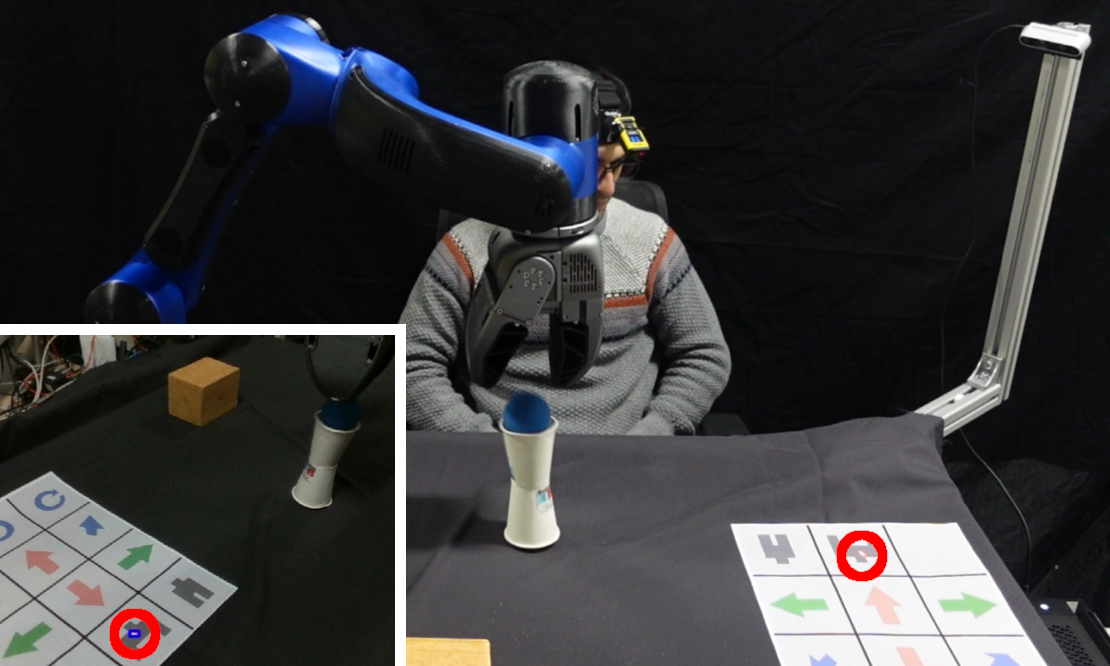}
	\vspace{-5px}	
	\caption{The \enquote{soft ball} pick-and-place experiment, executed by different users.}
	\vspace{-8px}
	\label{fig:exp3-frames}
\end{figure*}

\begin{figure}
	\centering
	\includegraphics[width=1\linewidth]{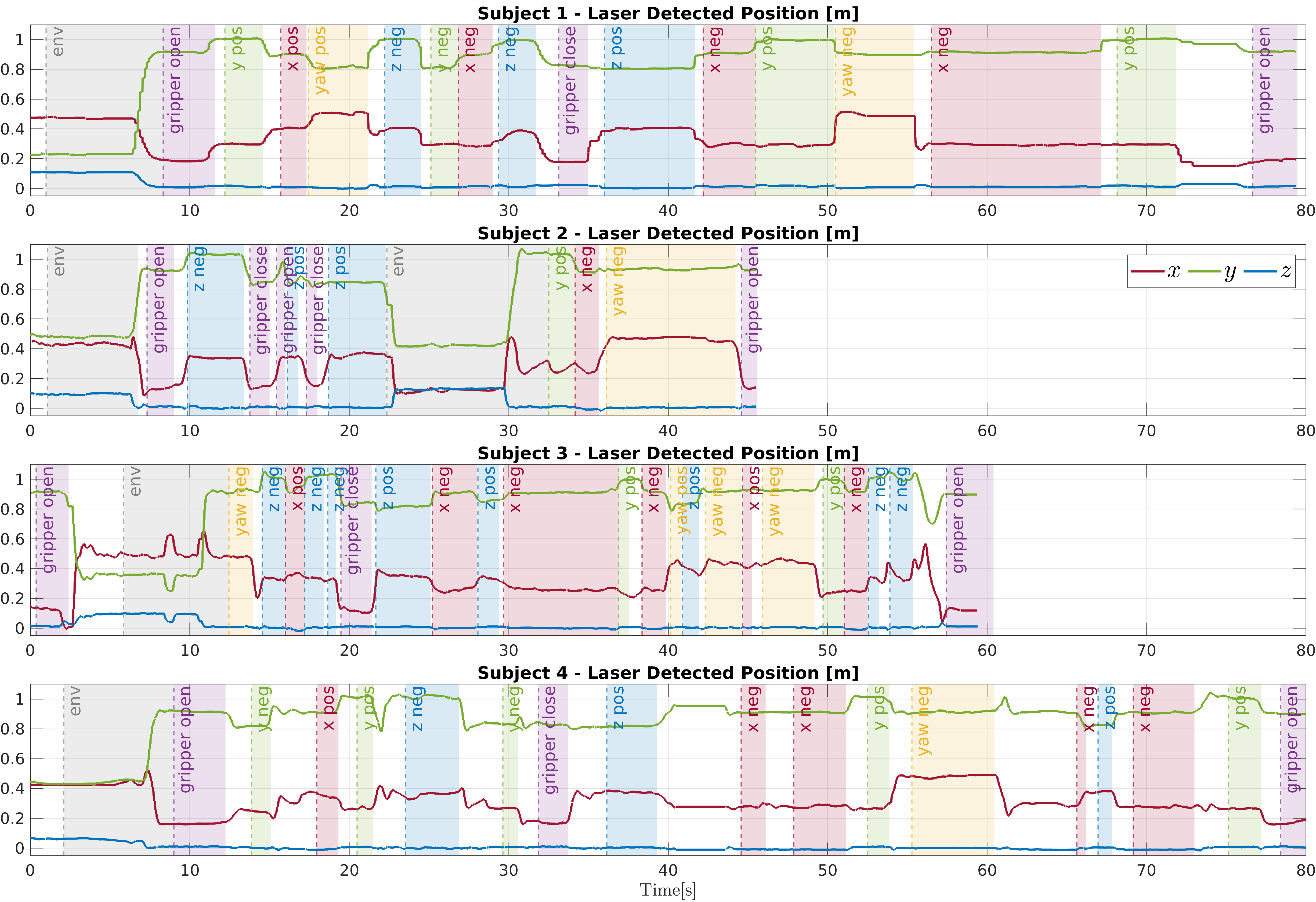}
	\vspace{-15px}
	\caption{\enquote{Wooden block} pick-and-place experiments plots. Each row represents the laser spot position in the task conducted by a specific subject. The activated robot commands are highlighted by the colored areas.}
	\vspace{-10px}
	\label{fig:exp2-all}
\end{figure}

\subsection{Collaboration and Pick-and-Place Experiments}
Two sets of experiments were considered, each one involving different kinds of upper limbs disabilities.
In the first set, we considered the case of individuals affected by stroke or other similar impairments that limit the motion capacity of a single arm, while the other arm is fully functional. In this scenario, the impaired arm is substituted by the robotic arm, which, guided by the user through the proposed interface, executes co-manipulation bimanual ADL tasks.
In the second set, we considered the case of people with impairments on both arms. Individuals with such disabilities often require assistance in more activities of daily living, including object transportation. In this scenario, we conducted pick-and-place demonstrations with different subjects, where the assistive robot executes all the actions as guided by the laser pointer worn by the users.
\figurename{}~\ref{fig:setup_with_frames} illustrates the main elements involved during the execution of a task with the system. On the left, the user is directing the laser to command the robot to reach the object indicated by the laser spot. The laser spot is detected by the vision system, as shown in the bottom-right window. In top-right area, the RViz window displays the detected laser spot position, the two manually added bounding boxes around the user and the table, and the octomap generated by MoveIt from the RGB-D camera data. These obstacles are taken into consideration in the generation of the collision-free trajectory toward the detected laser spot.

The first experiment, \enquote{cutting bread}, is shown in \figurename{}~\ref{fig:pane3-frames}. At the beginning, the user commands the robot to reach the bread by directing the head at it which results in pointing the laser on it. Subsequently, by pointing at the paper keyboard, the user is able to control the robot end-effector commanding it to hold the bread. With the bread steadily held in the gripper, the user proceeds to cut it using the healthy arm. The corresponding plots are shown in \figurename{}~\ref{fig:pane3-plot-all}. The first row displays the detected laser position with respect to the reference frame \textit{base\_link}. The second and third rows show the Cartesian pose (position and orientation) of the end-effector, while the fourth and fifth rows the Cartesian velocity (linear and angular) of the end-effector, referring as before to the \textit{base\_link} frame. These rows also show the reference Cartesian pose (dotted lines) which represents the target goal pose when operating in the environment control mode.
The last row presents the state of the gripper, showing its joint position (with position $0$ corresponding to the closed gripper) and effort.
The colored areas represent the time intervals during which the user is commanding the robot, corresponding either to the use of the environment control mode or to the activation of a specific button of the paper keyboard. All the commands represented by the colored areas are displayed in the top row plot. In the other plots, only the relevant commands are shown (e.g.\ gripper actions buttons are displayed in the plots relative to the gripper state). 
The laser position plot, at the end of the experiment, shows no data because the laser is outside the camera view since the user does not need to point anything while he is cutting the bread. It is also worth noticing that the gripper effort value in this time interval exhibits chatter, attributed to the user interaction with the bread during the cutting phase.

In the second experiment, illustrated in \figurename{}~\ref{fig:bottiglia-frames}, the user collaborates with the robot to open and close a bottle. Initially, the user points the laser at the bottle to command the robot to reach it. Subsequently, he employs the paper keyboard to grasp the bottle from the cap and bring it closer. With the robot holding the bottle from the cap, the user uses his healthy arm to grasp the body of the bottle while simultaneously commanding an end-effector yaw rotation pointing the laser at the correspondent keyboard button to unscrew the cap. After drinking, the subject brings the bottle back to the robot, commands an opposite end-effector yaw rotation to screw the cap, and then move the robot to place the bottle back on the table. Plots relevant to this experiment are shown in \figurename{}~\ref{fig:bottiglia-plot-all}, using the same layout as the plots of the previous experiment. In the laser position plot, the interval around $\SI{60}{\second}$ shows no data because the laser is outside the camera view, which happens while the user is drinking. 

\begin{figure}
	\centering
	\includegraphics[width=1\linewidth]{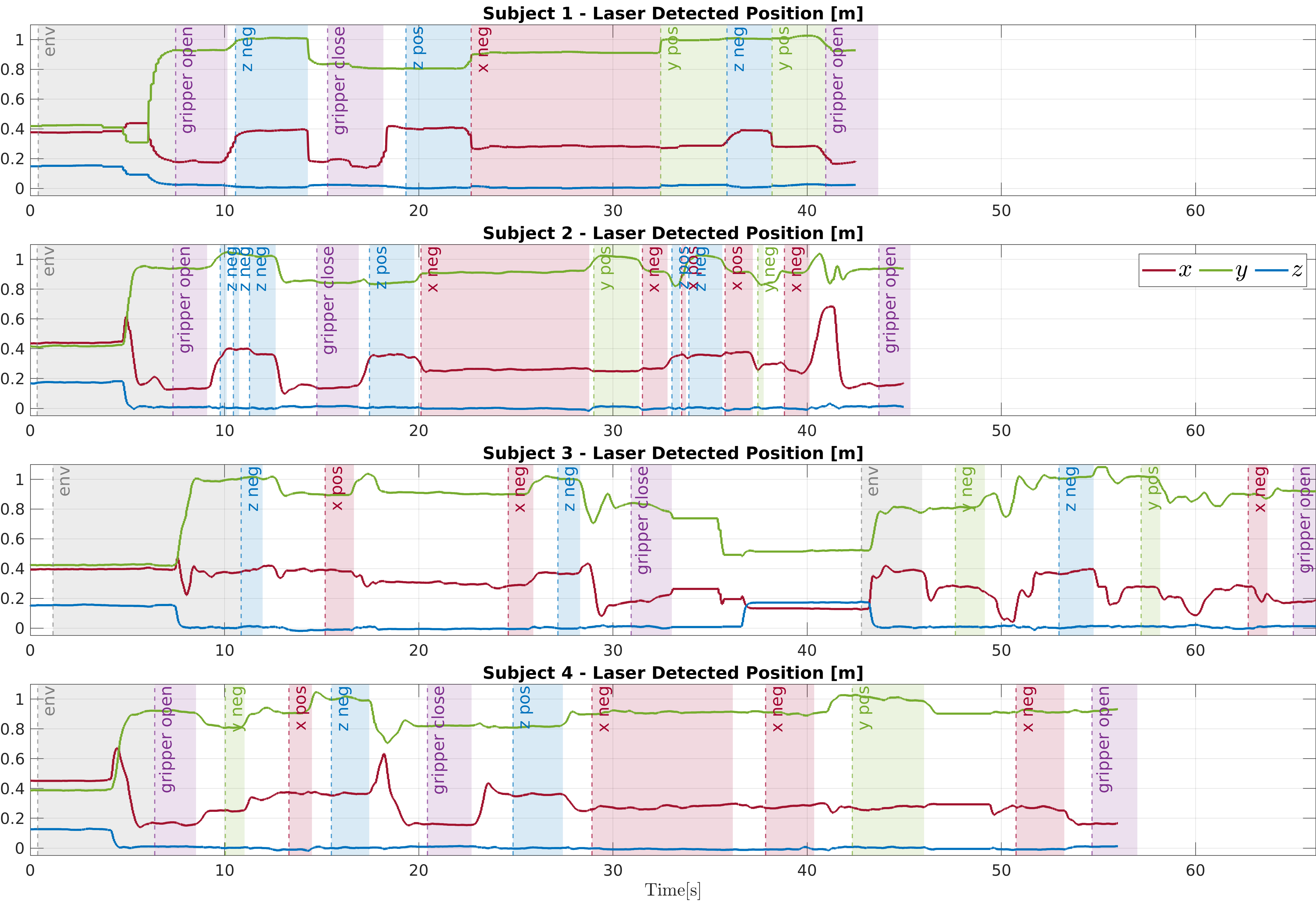}
	\vspace{-15px}
	\caption{\enquote{Soft ball} pick-and-place experiments plots. Each row represents the laser spot position in the task conducted by a specific subject. The activated robot commands are highlighted by the colored areas.}
	\vspace{-15px}
	\label{fig:exp3-all}
\end{figure}

In the other set of experiments users have no available upper limbs motions, hence they can not participate physically in the tasks. Therefore, they exclusively command the manipulator to accomplish the tasks with the interaction control modalities of the proposed interface.
The third experiment, shown in \figurename{}~\ref{fig:exp2-frames}, involves transporting a wooden block inside a container, while the fourth experiment, shown in \figurename{}~\ref{fig:exp3-frames} requires putting a soft ball inside a small cardboard glass. In the images, the four subjects involved in the experiments are shown in different moment of the tasks.
These two experiments share similar characteristics, but they have different kinds of challenges. In the first one, end-effector yaw rotations are necessary to align the gripper with the object to grasp and with the container. Instead, the second one demands more precision due to the small size of the ball and of the glass. 
The plots related to the wood block experiment and to the soft ball experiment are shown in \figurename{}~\ref{fig:exp2-all} and \figurename{}~\ref{fig:exp3-all}, respectively. Each row represents the laser position as pointed by the four subjects involved. Highlighted areas represent the command triggered by the laser spot position. 
In all the pick-and-place trials, users effectively combine the environment control mode and the keyboard control mode to grasp and transport the objects successfully into their respective containers. It can be observed how different choices have been made according to the user preferences. For example, Subject 2 with the wooden block and Subject 3 with the soft ball have employed the environment control mode also to command the robot toward the container, and not only at the beginning of the experiment to reach the objects demonstrating the intuitiveness of both control modes and the flexibility to combine them.
Furthermore, the availability of the two control modalities along with their effortless interchangeability allowed users to complete the tasks within a reasonable amount of time, as shown in the plots.

The experiments showcased the versatility of our interface, its effectiveness, and the intuitive control it provides to single and double-arm impaired users, enabling them to interact with the assistive robot arm and collaboratively execute ADL tasks. The demonstrations highlighted the efficacy of the system in intuitively controlling the robot using only head movements to direct the spot of the laser emitter. 
All the experiments described have been recorded and shown in the video accompanying this manuscript, available also at \url{https://youtu.be/WyWfgpezwRs}.

    \section{Conclusions}\label{sec:conclusions}

We have presented a novel laser-guided human-robot interaction interface to intuitively control an assistive robotic manipulator by projecting a laser onto locations in the robot workspace. 
The interface is designed to assist individuals with upper limbs disabilities. Thus, the laser emitter is conveniently worn on the head of the person allowing to guide the robot solely with head movements. The laser projection is detected by a neural-network vision pipeline without requiring to directly track the user head movements.
The implemented architecture incorporates two interchangeable control modalities. In the first modality, pointing the laser at a specific location commands the robot to move to that location, generating a collision-free trajectory. This feels very natural since users simply needs to direct their head toward the target, resulting in the laser to be pointed in the wanted location.
The second modality utilizes a paper keyboard, with buttons that can be virtually pressed by directing the laser onto them. The keyboard enables a more direct control of the robot, allowing the user to command end-effector Cartesian velocities and gripper actions. 
By integrating these two modalities in the proposed assistive laser-based interface, users can accomplish both co-manipulation and grasping tasks with the robot. Furthermore, they have the flexibility to choose the strategy that best fits their motion capabilities, preferences, and task needs.
We have conducted several experiments with an assistive manipulator to demonstrate the intuitiveness and effectiveness of the proposed interface considering scenarios of individuals with single and double-arm impairments, intentionally simulated by healthy subjects. 
In the future, we plan to develop and integrate more autonomous robot capabilities, such as autonomous planning and grasping of a selected object. 
Further experiments will involve users with the targeted disabilities, to conduct user studies and explore more the capability of the interface. Efforts will focus on investigating any specific residual motions that users may have in their impaired arm or arms to further extend the control functionalities of the introduced interface.
These advancements will contribute to the development of more sophisticated assistive robotic systems and their application in diverse home-care settings.

    \balance
	
	\bibliographystyle{IEEEtranBST/IEEEtran}
	\bibliography{IEEEtranBST/IEEEabrv,bib.bib}

\end{document}